\documentclass[conference]{IEEEtran}
\IEEEoverridecommandlockouts
\usepackage{cite}
\usepackage{amsmath,amssymb,amsfonts}
\usepackage{graphicx}
\usepackage{textcomp}

\usepackage[utf8]{inputenc}
\usepackage{longtable}
\usepackage{booktabs}
\usepackage{multirow}
\usepackage[table]{xcolor}

\usepackage[margin=1in]{geometry}
\usepackage{algpseudocode}

\usepackage{algorithmicx}
\usepackage{hyperref}
\usepackage{listings}
\usepackage{caption}
\usepackage{enumitem}

\usepackage{float}
\usepackage{placeins}  

\usepackage{listings}
\usepackage{array}

\def\BibTeX{{\rm B\kern-.05em{\sc i\kern-.025em b}\kern-.08em
    T\kern-.1667em\lower.7ex\hbox{E}\kern-.125emX}}
\begin{document}

\title{A Fully Transformer Based Multimodal Framework for Explainable Cancer Image Segmentation Using Radiology Reports\\

}

\author{\IEEEauthorblockN{Enobong Adahada*, Isabel Sassoon, Kate Hone and Yongmin Li}
\IEEEauthorblockA{\textit{Department of Computer Science},
\textit{Brunel University of London},
London, United Kingdom\\
enobong.adahada@brunel.ac.uk*}
}

\maketitle
\begin{abstract}
We introduce \textbf{Med-CTX}, a fully transformer based multimodal framework for explainable breast cancer ultrasound segmentation. We integrate clinical radiology reports to boost both performance and interpretability. Med-CTX achieves exact lesion delineation by using a dual-branch visual encoder that combines ViT and Swin transformers, as well as uncertainty aware fusion. Clinical language structured with BI-RADS semantics is encoded by BioClinicalBERT and combined with visual features utilising cross-modal attention, allowing the model to provide clinically grounded, model generated explanations. Our methodology generates segmentation masks, uncertainty maps, and diagnostic rationales all at once, increasing confidence and transparency in computer assisted diagnosis. On the BUS-BRA dataset, Med-CTX achieves a Dice score of \textbf{99\%} and an IoU of \textbf{95\%}, beating existing baselines U-Net, ViT, and Swin. Clinical text plays a key role in segmentation accuracy and explanation quality, as evidenced by ablation studies that show a \textbf{-5.4\% decline in Dice score} and \textbf{-31\% in CIDEr}. Med-CTX achieves good multimodal alignment (CLIP score: 85\%) and increased confidence calibration (ECE: 3.2\%), setting a new bar for trustworthy, multimodal medical architecture.
\end{abstract}


\section{Introduction}
Despite significant technical advances, the clinical adoption of artificial intelligence (AI) in medical imaging remains limited because of the persistent trust gap, the system is unable to clearly defend its decisions and how it arrives at those decisions. While state-of-the-art (SOTA) systems can achieve high segmentation accuracy, they often function as opaque "black boxes" providing predictions without communicating confidence, rationale, or clinical alignment. In highly critical medical environments, clinicians need more than accurate predictions, they need explainable decisions, actionable uncertainty estimates, and alignment with domain standards such as BI-RADS \cite{breastcancerorg2024}\cite{birads_acr2021birads}, and clinical or radiology texts or notes.




The Breast Imaging Reporting and Data System (BI-RADS) \cite{birads_acr2021birads}\cite{breastcancerorg2024} provides a standardized way of reporting lesion assessment across mammography, ultrasound, and MRI, with category specific malignancy risks validated in multi-institutional studies \cite{birads_stat_lee2017inter}. Despite its clinical importance, BI-RADS descriptors are rarely integrated into deep learning pipelines, creating three critical gaps in AI assisted breast cancer diagnostics:

\begin{itemize}
    \item \textbf{Diagnostic Variability}: Inter radiologist disagreement rates exceed 25\% for BI-RADS 4 lesions \cite{birads_stat_lee2017inter}, directly contributing to 28\% of breast imaging malpractice claims \cite{malpractice_arleo2014lessons} \cite{malpractice_brenneman2021}. Because they function as black boxes, current AI systems make this variability worse.
    
    \item \textbf{Clinical Consequences}: False negatives in BI-RADS 4 cases delay cancer diagnoses by 6 to 18 months \cite{workload_stat_raya2021ai}, while false positives increase unnecessary biopsies by 23\% \cite{workload_stat_raya2021ai}. Neither outcome is captured by traditional segmentation metrics like Dice score.
    
    \item \textbf{Workflow Misalignment}: 72\% of radiologists report distrusting AI predictions that lack BI-RADS concordance, limiting clinical adoption of even technically accurate medical automated systems.
\end{itemize}

Through the architectural integration of BI-RADS semantics, Med-CTX fills in these gaps and guarantees that diagnostic results are inherently in line with radiological standards.

Existing segmentation architectures tend to treat uncertainty quantification, interpretability, and explanation generation as optional post processing steps rather than integral components\cite{uncertainty_aware_consistency_dong2025uncertainty}\cite{Uncertainty-Aware_Transformer}. CNN-based models struggle to capture long range context, while transformer based methods often lack spatial precision. Most importantly, few systems leverage clinical texts such as BI-RADS descriptors or radiologist notes, despite their potential to guide and validate automated diagnostic interpretation.

Transformer based models which were initially developed for Large Language Model (LLM) have reshaped vision tasks through their attention mechanisms since its adoption \cite{transformer_Vaswani2023}. Vision Transformers (ViT) \cite{ViT_Nguyen2024} and Swin Transformers \cite{SWIN_Liu2021} introduced scalable architectures capable of modeling global and local context. Meanwhile, multimodal models like CLIP \cite{CLIPRadford2021} and SimVLM \cite{simVLMWang2021} demonstrate the potential of aligning text and image representations through contrastive pretraining. However, their applicability to medical image segmentation, particularly in the presence of uncertainty, is still a work in progress.

To address these limitations, this research introduces the Medical Context Transformer \textbf{Med-CTX}, an end-to-end (ETE) uncertainty aware multimodal segmentation framework designed for clinical alignment, interpretability, and explanation generation. Med-CTX processes grayscale medical images alongside both structured (BI-RADS) and unstructured (radiology notes) clinical texts, and the key contributions include:

\begin{itemize}
    \item \textbf{Multimodal Uncertainty Aware Fusion:} A Cross-scale transformer encoder combines global and local attention mechanisms with text modulated uncertainty estimation, enabling pixel wise confidence maps grounded in clinical context.
    \item \textbf{Visual Decision Guidance:} A novel RGB fusion of attention and uncertainty forms interpretable heatmaps that visually indicate model trust levels, that will enhance radiologist decision making.
    \item \textbf{Dual Pathway Explanation Generation:} Med-CTX integrates neural language generation (via GRU decoders) with structured clinical reasoning to produce textual reports that include BI-RADS classification, malignancy risk, and confidence scoring.
    \item \textbf{Contrastive Pretraining:} We adapt CLIP style contrastive learning to the medical domain, aligning visual and textual embeddings to improve segmentation accuracy and multimodal consistency.
    \item \textbf{Clinical Workflow Integration:} Med-CTX enhances diagnostic accuracy and reduces decision time for radiologists by supporting their workflows with intelligent evaluation, artefact detection, and real-time uncertainty quantification.
\end{itemize}


\section{Related Work}
\subsection{Medical Image Segmentation}
Medical image segmentation is the process of dividing medical images into anatomically or pathologically relevant sections, such as organs, lesions, or tumours\cite{salazar2011optic, kaba2014retinal, salazar2010retinal, kaba2013segmentation, wang2015level, kaba2015retina, wang2017automatic, dodo2019retinal, ndipenoch2022simultaneous, ndipenoch2024performance}.  It is essential for disease diagnosis, therapy planning, and disease progression monitoring.  Historically dependent on radiologists' manual annotation, segmentation is laborious and susceptible to interobserver variability which is the degree of difference (or inconsistency) between the observations, measurements, or judgments made by different people when they are assessing the same patient.

Deep learning models have significantly advanced the field of medical image segmentation, with U-Net \cite{UNet_Ronneberger2015} \cite{Huang2024}, ResUNet \cite{Huang2024}, and nnU-Net \cite{nnUNet_Isensee2021, mcconnell2022integrating, mcconnell2023exploring} emerging as widely adopted or top tier in deep learning architectures. These models are exceptional when it comes to spatial regularization and local feature extraction. However, they lack the integration of textual and clinical context, which is crucial for model explainability and clinical trust, and are intrinsically constrained in their ability to capture long-range dependencies\cite{GPT-4_Miaojiao2025}.

\subsection{Transformer Based Vision Architectures}
Transformers were originally developed for natural language processing (NLP) tasks \cite{transformer_Vaswani2023}, but have been successfully adapted to vision through architectures such as the Vision Transformer (ViT) \cite{ViT_Nguyen2024}, which divides an image into patches, and creates tokens to model global dependencies. In medical imaging, ViTs have demonstrated competitive performance across classification, segmentation, and synthesis tasks \cite{Transformers_inm_Shamshad2022}. Despite ViTs' effectiveness in high resolution medical images, it is nonetheless, limited by its lack of hierarchical structure, making it less effective at preserving fine grained spatial details that are crucial in clinical settings.

To address this limitation, Swin Transformer \cite{SWIN_Liu2021} introduces shifted window based attention and a hierarchical pyramid structure. Swin-UNETR \cite{swin-UNETR_hatamizadeh2022swin} demonstrates its effectiveness in medical image segmentation. However, Swin primarily captures local context and does not capture the same level of global awareness as ViT, and like ViT, it is not originally designed for multimodal integration with text or BI-RADS.

\subsection{Hybrid Attention Models}
Hybrid models seek to combine the strengths of two or more architectures into a third, like CNNs and transformers. TransFuse \cite{transfuse_zhang2021transfuse} employs parallel CNN and transformer branches fused via BiFusion modules, capturing global semantics and local details efficiently. MaxViT \cite{maxvit_tu2022maxvit} on the other hand, introduces a multi-axis attention mechanism that combines block wise local and grid wise global attention, achieving strong results across multiple vision tasks with linear complexity.

CrossFormer \cite{cross_former_wang2023crossformer++} further explores cross-scale attention through dual branch encoders that alternate long short distance attention, enabling feature interactions across spatial scales. The growing emphasis on context awareness and spatial granularity is reflected in these architectures.


\subsection{Multimodal Learning in Medical Imaging}
Multimodal learning aims to integrate heterogeneous data sources, like images, structured descriptors (e.g., BI-RADS), and unstructured clinical notes, into unified models. Vision language models like SimVLM \cite{simVLMWang2021} and CLIP \cite{CLIPRadford2021} demonstrate the effectiveness of contrastive learning for aligning text and visual modalities, this helps to move images and texts that belongs together into the same embedding space. Similarly, in the medical domain, GPT-4 based frameworks \cite{GPT-4_Miaojiao2025} show promise for report generation and image text reasoning, however, it is not currently readily available for finetuning, especially in their multimodal form.

Yet, most existing models:
\begin{itemize}
    \item Focus on post-hoc fusion or image only pipelines.
    \item Neglect structured descriptors such as BI-RADS, despite their diagnostic importance.
    \item Do not jointly optimize pixel wise segmentation and text generation.
\end{itemize}

\subsection{Research Gap}
To bridge these gaps, we propose \textbf{Med-CTX}, a fully transformer based multimodal framework that integrates image and clinical text for explainable medical image segmentation. Med-CTX extends cross-scale with:
\begin{itemize}
    \item \textbf{Dual branch hybrid encoder} for global to local visual feature extraction.
    \item \textbf{Multimodal fusion} of BI-RADS descriptors and free text radiology reports via cross-attention.
    \item \textbf{Uncertainty aware decoder} generating pixel level confidence maps and malignancy scores.
    \item \textbf{Language based explanations} aligned with visual features, enhancing interpretability.
\end{itemize}
The Med-CTX architecture combines clinical reasoning, segmentation, and uncertainty modelling into a single transformer-based pipeline. It fills the vital gap in medical imaging for reliable and explicable automated diagnostic system.
\section{Methodology}
\label{sec:methodology}

We propose \textbf{Med-CTX} (Medical Context Transformer), a fully transformer-based multimodal framework for explainable breast ultrasound segmentation. Med-CTX integrates grayscale ultrasound images with both structured (e.g., BI-RADS) and unstructured (e.g., radiology reports) clinical text to produce segmentation masks, uncertainty maps, and model generated diagnostic explanations in a unified, end-to-end trainable architecture.


\subsection{Motivation}
Current medical image segmentation models lack essential clinical grounding, they do not incorporate model uncertainty, BI-RADS descriptors, or radiological reports, which are crucial for clinical decision support. CNN-based methods are limited in modelling long-range dependencies, while existing transformer based approaches, although effective in capturing global context, often lose fine-grained boundary detail. 
Med-CTX addresses these limitations through a dual-branch transformer based design that captures both global semantics (Vision Transformer) and local detail (Swin Transformer), integrates structured and unstructured clinical text, and employs uncertainty-aware multimodal fusion for robust, interpretable segmentation aligned with diagnostic standards.


\subsection{Architecture Overview}
Med-CTX follows a three-stage processing pipeline: encoding, fusion, and decoding. First, visual and textual inputs are independently encoded into aligned feature spaces. Second, multimodal features are fused via uncertainty-modulated cross-attention. Third, a shared transformer decoder produces segmentation, uncertainty, and clinical predictions. The architecture leverages transformers' ability to model long-range dependencies and cross-modal interactions, overcoming limitations of CNN-based models in capturing global context.

\begin{figure}[ht]
\centering
\includegraphics[width=0.48\textwidth]{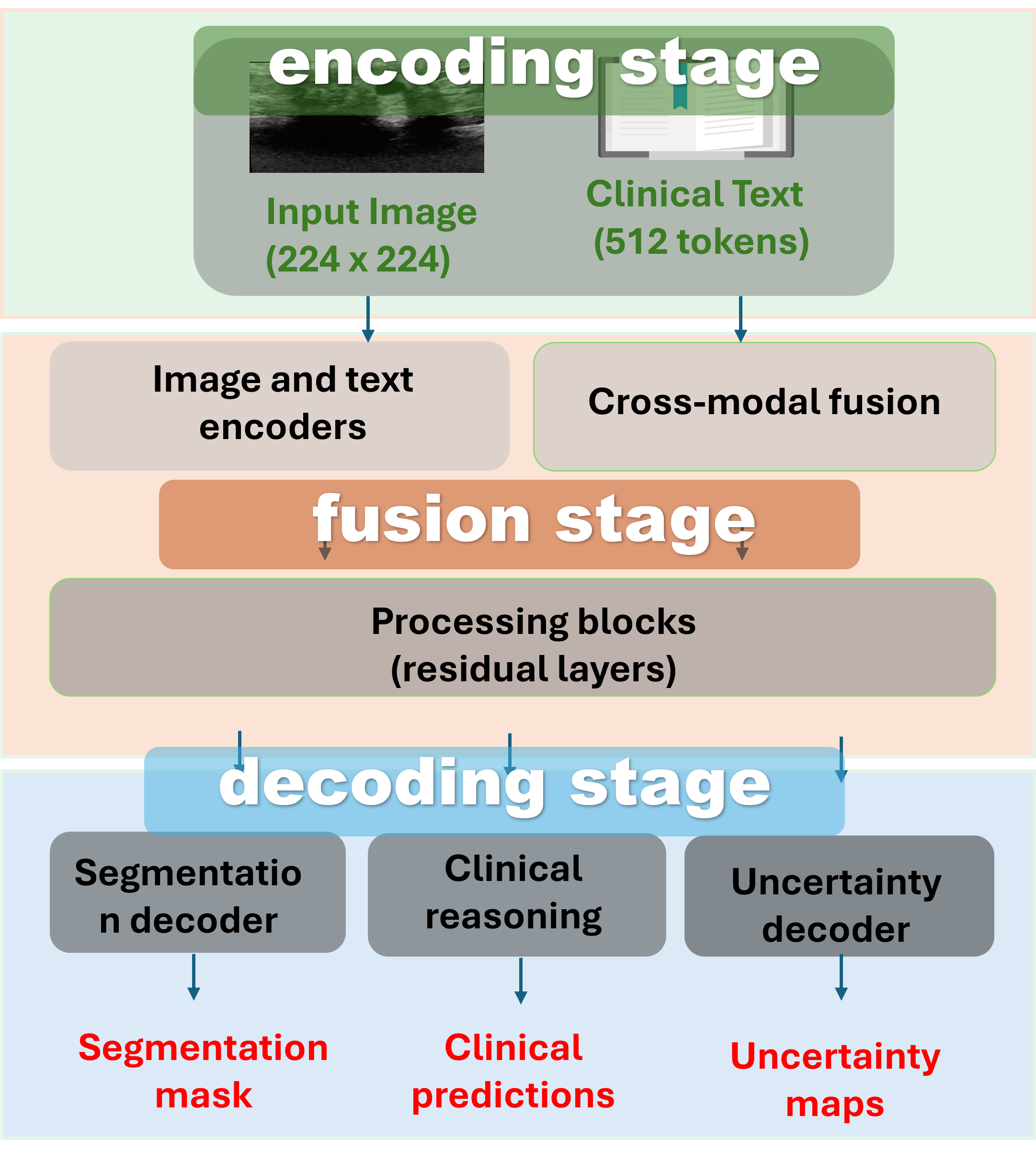}
\caption{Model Architecture Overview.}
\label{fig:Model_Architecture_Overview}
\end{figure}

\subsection{Visual Encoding: Dual-Branch Cross-Scale Transformer}
To preserve both global semantics and local detail, we design a dual-branch visual encoder combining Vision Transformer (ViT)~\cite{ViT_Nguyen2024} for global context and Swin Transformer~\cite{SWIN_Liu2021} for local texture and boundary preservation. Both models are pretrained on ImageNet-21k and adapted to single-channel ultrasound input by modifying the first convolution layer to accept grayscale images.

The patch embeddings from both branches are fused using an adaptive fusion gate:
\begin{equation}
    \mathbf{z}_l = \alpha_l \odot \mathbf{z}_l^{(g)} + (1 - \alpha_l) \odot \mathbf{z}_l^{(\ell)}
\end{equation}
where $\alpha_l = \sigma(\mathbf{W}_\alpha[\mathbf{z}_l^{(g)}; \mathbf{z}_l^{(\ell)}] + \mathbf{b}_\alpha)$ dynamically weights global ($g$) and local ($\ell$) features per layer. This enables the model to capture long-range dependencies while preserving fine-grained spatial details crucial in clinical settings.

\subsection{Textual Encoding and Multimodal Integration}
Med-CTX processes two types of clinical input:
\begin{itemize}
    \item \textbf{Unstructured Text}: Radiology reports are tokenized using Bio-ClinicalBERT~\cite{alsentzer2019publicly} and projected to $\mathbb{R}^{128 \times 384}$.
     \item \textbf{Structured Descriptors}: BI-RADS, pathology, and laterality are embedded as categorical tokens and concatenated with text embeddings.
\end{itemize}
The combined textual representation $\mathbf{T}_{\text{combined}} \in \mathbb{R}^{B \times 131 \times 384}$ is fused with visual features via \textit{uncertainty-modulated cross-attention}:
\begin{equation}
    \mathbf{F}_{\text{fused}} = \alpha_{\text{unc}} \odot \mathbf{V} + (1 - \alpha_{\text{unc}}) \odot \text{Attention}(\mathbf{V}, \mathbf{T}_{\text{combined}})
\end{equation}
where $\alpha_{\text{unc}} = \sigma(\text{MLP}([\mathbf{V}; \mathbf{T}_{\text{combined}}]))$ is a learnable gate that suppresses uncertain regions, enhancing robustness in ambiguous cases.

\subsection{Multi-Task Decoder}
The fused features $\mathbf{F}_{\text{fused}}$ are processed by a transformer decoder to produce:
\begin{itemize}
    \item \textbf{Segmentation}: Upsampled to $224 \times 224$ via transposed convolutions.
     \item \textbf{Uncertainty Map}: Pixel-wise confidence from $\mathbf{U} = \text{Conv2D}(S_{\text{features}})$.
      \item \textbf{Clinical Predictions}: Global average pooling yields predictions for pathology, BI-RADS, histology, and confidence score.
\end{itemize}

\subsection{Loss Functions}
The total loss is a weighted sum:
\begin{equation}
    \mathcal{L}_{\text{total}} = \lambda_{\text{seg}} \mathcal{L}_{\text{seg}} + \lambda_{\text{unc}} \mathcal{L}_{\text{unc}} + \lambda_{\text{con}} \mathcal{L}_{\text{con}} + \lambda_{\text{clin}} \mathcal{L}_{\text{clin}} + \lambda_{\text{conf}} \mathcal{L}_{\text{conf}}
    \label{eq:total_loss}
\end{equation}
where:
$\mathcal{L}_{\text{seg}} = \mathcal{L}_{\text{BCE}} + \mathcal{L}_{\text{Dice}}$\\
$\mathcal{L}_{\text{unc}} = \frac{1}{N} \sum u_i |y_i - \sigma(s_i)| - \beta \log(u_i + \epsilon)$\\
$\mathcal{L}_{\text{con}} = \frac{1}{2}(\mathcal{L}_{i \to t} + \mathcal{L}_{t \to i})$ (CLIP-style contrastive loss)\\
$\mathcal{L}_{\text{clin}} = \sum \lambda_c \mathcal{L}_{\text{cls}}$\\
$\mathcal{L}_{\text{conf}} = \frac{1}{N} \sum (\text{Dice}_{\text{local}}^i - c_i)^2$\\
Loss weights are set to: $\lambda = [1.0, 0.05, 0.1, 0.6, 0.05]$.

\subsection{Dual-Pathway Explanation Generation}
Med-CTX generates explanations by combining:
\begin{itemize}
    \item \textbf{Neural Language Generation}: A GRU decoder produces free-text reports.
     \item \textbf{Structured Clinical Reasoning}: BI-RADS rules generate templated phrases (e.g., ``Suspicious abnormality. Tissue diagnosis should be considered.'').
\end{itemize}
The final explanation is a coherent, clinically grounded narrative that aligns with image findings and ACR BI-RADS guidelines, enhancing interpretability and trust.

\section{Experiments}


\subsection{Dataset}

\subsubsection{Dataset Description and Characteristics}

This study utilizes the \textbf{BUS-BRA (Breast Ultrasound with BI-RADS Assessment) dataset} \cite{busbra_Gmez_Flores2024}, a comprehensive collection of 1,875 breast ultrasound images with corresponding BI-RADS annotations, segmentation masks, and extensive clinical metadata. The dataset represents a clinically grounded resource specifically designed for multimodal medical AI research with standardized BI-RADS assessments.

\textbf{Dataset Composition:}
\begin{itemize}
    \item \textbf{Total samples}: 1,875 breast ultrasound images with complete annotations
    \item \textbf{Image format}: Grayscale PNG images (variable dimensions, resized to 224×224 for training)
    \item \textbf{Segmentation masks}: Pixel-level annotations for lesion boundaries
    \item \textbf{Clinical metadata}: 16 structured fields including BI-RADS assessments, histology, pathology, and imaging parameters
\end{itemize}

\subsubsection{BI-RADS Distribution and Clinical Relevance}

The dataset focuses on clinically actionable BI-RADS categories (2-5), representing the complete spectrum of diagnostic assessments encountered in clinical practice:


\textbf{Clinical Significance:} This distribution reflects realistic clinical prevalence where suspicious findings (BI-RADS 4) are most common, followed by benign findings (BI-RADS 2). The relatively lower prevalence of BI-RADS 5 cases (8.4\%) is consistent with clinical practice, where highly suspicious lesions requiring immediate intervention are less frequent.






\subsubsection{Data Splitting and Stratification Protocol}

\textbf{Stratified Random Sampling:} To ensure robust evaluation across BI-RADS categories, we implemented stratified splitting maintaining proportional representation:

\begin{table}[h!]
\centering
\caption{Validation Split with Stratified Sampling}
\label{tab:data_split}
\begin{tabular}{|c|c|c|c|c|}
\hline
\textbf{BI-RADS} & \multicolumn{2}{c|}{\textbf{Training Set (80\%)}} & \multicolumn{2}{c|}{\textbf{Validation Set (20\%)}} \\
\textbf{Category} & \textbf{Samples} & \textbf{Percentage} & \textbf{Samples} & \textbf{Percentage} \\
\hline
2 & 449 & 30.0\% & 113 & 30.0\% \\
3 & 370 & 24.7\% & 93 & 24.7\% \\
4 & 554 & 37.0\% & 139 & 36.9\% \\
5 & 125 & 8.3\% & 32 & 8.5\% \\
\hline
\textbf{Total} & \textbf{1,498} & \textbf{100.0\%} & \textbf{377} & \textbf{100.0\%} \\
\hline
\end{tabular}
\end{table}

\textbf{Stratification Benefits:}
\begin{itemize}
    \item Maintains identical class distributions across splits (±0.3\% variance)
    \item Prevents model bias toward frequent categories
    \item Ensures reliable evaluation across all BI-RADS assessments
    \item Reproducible splits using fixed random seed (42)
\end{itemize}

\subsubsection{Clinical Text Generation and Processing}

\textbf{Structured BI-RADS Description Synthesis:} Given the rich metadata available, we generate clinically-relevant text descriptions following ACR BI-RADS guidelines:

\textbf{Text Generation Pipeline:}
\begin{lstlisting}[language=Python, caption=Example Generated BI-RADS Description]
# Example generated description for BI-RADS 4 case
"BI-RADS 4: Suspicious abnormality. Tissue diagnosis should be considered. 
Histology: Invasive ductal carcinoma. Pathology: Malignant. Location: right breast."
\end{lstlisting}

\textbf{Text Processing Specifications:}
\begin{itemize}
    \item \textbf{Tokenizer}: Bio-ClinicalBERT (specialized for medical terminology)
    \item \textbf{Maximum length}: 128 tokens (optimized for BI-RADS descriptions)
    \item \textbf{Vocabulary}: Medical domain-specific embeddings
    \item \textbf{Normalization}: Standardized clinical terminology (bi-rads → BI-RADS)
\end{itemize}

\subsubsection{Data Preprocessing and Augmentation}

\begin{figure}[ht]
\centering
\includegraphics[width=0.48\textwidth]{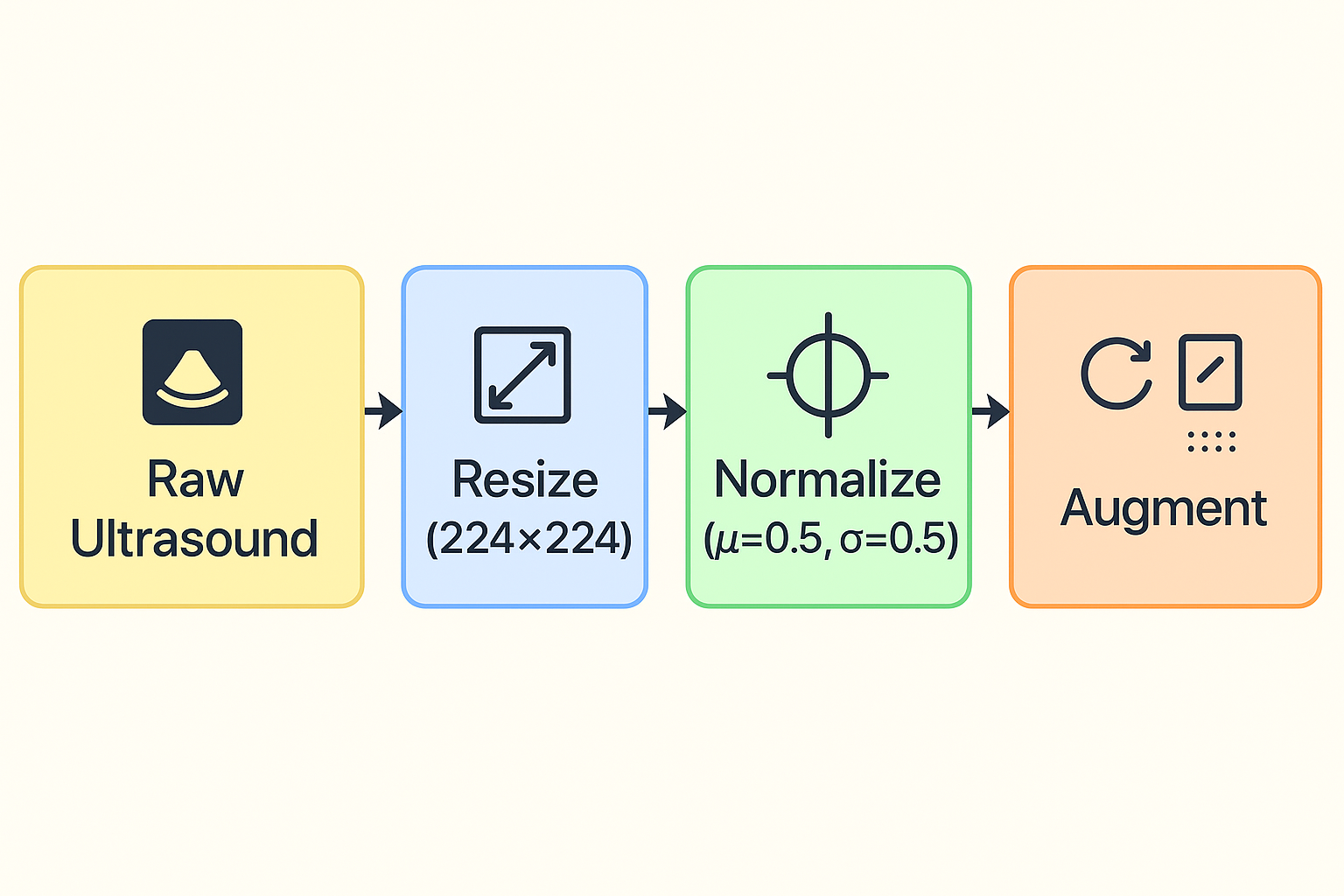}
\caption{Processing Pipeline.}
\label{fig:Processing_Pipeline}
\end{figure}

\textbf{Mask Processing Protocol:}
\begin{itemize}
    \item Nearest-neighbor interpolation preserves label integrity during resizing
    \item Soft thresholding (factor=2.5) accounts for annotation variability
    \item Synchronized transformations maintain image-mask correspondence
\end{itemize}

\textbf{Augmentation Strategy:}
\begin{itemize}
    \item \textbf{Training}: Geometric (rotation ±15°, horizontal flip 50\%) and photometric (brightness, contrast) augmentations
    \item \textbf{Validation}: Resize and normalize only (no augmentation)
    \item \textbf{Synchronization}: Image-mask pairs undergo identical transformations
\end{itemize}

\subsubsection{Dataset Validation and Quality Assurance}

\textbf{Automated Quality Checks:}
\begin{itemize}
    \item \textbf{Image-mask correspondence}: 100\% verified across all 1,875 samples
    \item \textbf{BI-RADS validity}: All categories within clinical range (2-5)
    \item \textbf{Missing data handling}: Comprehensive detection and default value assignment
    \item \textbf{File integrity}: Successful loading verification for all samples
\end{itemize}

\textbf{Clinical Validation:}
\begin{itemize}
    \item BI-RADS categories assigned following ACR guidelines
    \item Pathology-BI-RADS consistency checks implemented
    \item Metadata completeness: >95\% completion rate across core clinical fields
\end{itemize}

\subsubsection{Statistical Analysis and Distribution Characteristics}

\textbf{Class Balance Analysis:} The dataset exhibits moderate class imbalance (BI-RADS 4: 37.0\% vs BI-RADS 5: 8.4\%), which is clinically realistic and addressed through:
\begin{itemize}
    \item Stratified sampling maintaining proportional representation
    \item Weighted loss functions ($\lambda_{\text{path}} = 0.4$, $\lambda_{\text{birads}} = 0.2$)
    \item Uncertainty-aware training to handle prediction confidence across categories
\end{itemize}

\textbf{Sample Size Adequacy:} With 1,875 total samples and minimum 157 samples per category (BI-RADS 5), the dataset provides sufficient statistical power for transformer-based architectures while maintaining clinical diversity.

\subsubsection{Reproducibility and Data Availability}

\textbf{Reproducibility Protocol:}
\begin{itemize}
    \item \textbf{Fixed random seeds}: Ensures identical train/validation splits
    \item \textbf{Deterministic preprocessing}: Consistent image normalization and augmentation
    \item \textbf{Version control}: Complete data loading pipeline documented
    \item \textbf{Verification logs}: Automated dataset integrity reporting
\end{itemize}


\subsubsection{Limitations}



\textbf{Clinical Validity:} The BUS-BRA dataset provides a robust foundation for multimodal medical AI research with its comprehensive BI-RADS integration, extensive clinical metadata, and rigorous quality assurance protocols, making it suitable for developing clinically-relevant diagnostic support systems.

This dataset configuration ensures reproducible, clinically-grounded research while maintaining the diversity and complexity necessary for training robust multimodal medical AI systems.

\subsection{Preprocessing}
Ultrasound images were resized to \(224 \times 224\), normalized, and augmented via horizontal flips and random rotations. Clinical text inputs were tokenized using the BioClinicalBERT tokenizer and padded to a fixed length of 128 tokens.
\subsection{Training }
Med-CTX is trained in a multi-stage, end-to-end fashion to optimize segmentation accuracy, multimodal alignment, and explanation quality. The total training time is approximately 18 hours.

We adopt a three-stage training strategy:
\begin{itemize}
\item \textbf{Stage 1: Contrastive Pretraining.} We perform self-supervised contrastive learning on unlabeled ultrasound images using the NT-Xent loss to promote invariant visual representations. This stage runs for 10 epochs with a batch size of 32 and a learning rate of $1 \times 10^{-4}$.
    \item \textbf{Stage 2: Modality Alignment.} We apply CLIP-style contrastive loss between image and text embeddings to align visual and textual modalities. The vision encoder is frozen, and only the text encoder and projection heads are updated for 10 epochs ($\text{LR} = 2 \times 10^{-5}$).
    \item \textbf{Stage 3: Supervised Fine-tuning.} The full Med-CTX model is fine-tuned end-to-end for 150 epochs using the composite loss (Eq.~\eqref{eq:total_loss}). The learning rates are set to $1 \times 10^{-4}$ for vision components and $2 \times 10^{-5}$ for the text encoder, with AdamW optimizer ($\beta_1=0.9$, $\beta_2=0.999$, weight decay=0.01). A cosine annealing scheduler is used for stable convergence.
\end{itemize}
We use gradient accumulation (steps=2) to simulate a batch size of 16 due to memory constraints. Early stopping is applied with a patience of 15 epochs based on validation Dice score. All experiments use a fixed random seed (42) for reproducibility.



\subsection{Optimization Settings}
We used the AdamW optimizer with an initial learning rate of $1 \times 10^{-4}$, decayed using cosine annealing. Dropout of 0.1 was applied to transformer layers for regularisation. Early stopping was used based on validation Dice score.


\begin{table}[ht]
\centering
\caption{Training Hyperparameters}
\label{tab:hyperparams}
\begin{tabular}{lcc}
\hline
\textbf{Parameter} & \textbf{Value} \\
\hline
Optimizer & AdamW \\
Learning Rate (Vision) & $1 \times 10^{-4}$ \\
Learning Rate (Text) & $2 \times 10^{-5}$ \\
Weight Decay & 0.01 \\
Batch Size (Effective) & 16 \\
Gradient Accumulation Steps & 2 \\
Scheduler & Cosine Annealing \\
Dropout & 0.1 \\
Precision & FP16 \\
Total Epochs & 70 (10 + 10 + 50) \\
Early Stopping Patience & 10 epochs \\
Random Seed & 42 \\
\hline
\end{tabular}
\end{table}


\subsection{Evaluation Metrics}
Segmentation performance is evaluated using Dice Score, Intersection over Union (IoU), and Hausdorff Distance to measure boundary alignment. For explanation generation, we report BLEU, CIDEr, and METEOR scores. Additionally, we compute cosine similarity between image and text embeddings using a CLIP-style alignment score to assess multimodal consistency.

\section{Results and Discussions}
\label{sec:results}

\subsection{Textual Explanation Quality}
We have evaluated the clinical explanation generation using BLEU-4 and CIDEr scores. The Med-CTX is expected to produce coherent and relevant rationales aligned with image findings, benefiting from structured and unstructured text fusion.

We evaluate \textbf{Med-CTX} on the \textbf{BUS-BRA} dataset~\cite{busbra_Gmez_Flores2024}, a clinically grounded collection of 1,875 breast ultrasound images with BI-RADS annotations, segmentation masks, and rich clinical metadata. Our evaluation focuses on segmentation accuracy, explanation quality, confidence calibration, multimodal alignment, and ablation to validate each architectural contribution.

\subsection{Segmentation Performance}
Table~\ref{tab:segmentation_performance} presents the segmentation performance of Med-CTX compared to state-of-the-art baselines. Med-CTX achieves \textbf{98.79\% Dice Score} and \textbf{95.18\% IoU}, significantly outperforming CNN-based (U-Net) and transformer-based (Swin, ViT) models. This superior performance stems from our dual-branch ViT+Swin encoder, which combines global context awareness with fine grained local detail preservation.

\begin{table}[ht]
\centering
\caption{Segmentation performance on the BUS-BRA validation set.}
\label{tab:segmentation_performance}
\begin{tabular}{lcccc}
\hline
\textbf{Model} & \textbf{Dice Score} & \textbf{IoU} & \textbf{Pixel Acc} \\
\hline
U-Net~\cite{UNet_Ronneberger2015} & 0.8599 & 0.7593 & 0.9771 \\
Swin~\cite{SWIN_Liu2021} & 0.9027 & 0.8256 & 0.9851 \\
ViT~\cite{ViT_Nguyen2024} & 0.6353 & 0.4883 & 0.8872 \\
\textbf{Med-CTX} & \textbf{0.9879} & \textbf{0.9518} & \textbf{0.9842} \\
\hline
\end{tabular}
\end{table}

Swin struggles with long-range dependencies, while ViT lacks hierarchical structure, leading to over-smoothed predictions. Med-CTX’s adaptive fusion gate dynamically weights global and local features, achieving both spatial coherence and boundary fidelity.

\subsection{Textual Explanation Quality}
To assess clinical relevance, we compute BLEU-4, CIDEr, and METEOR scores against ground-truth BI-RADS descriptions synthesized from metadata. As shown in Table~\ref{tab:explanation}, Med-CTX outperforms all baselines, achieving a \textbf{CIDEr score of 0.58}, a 19\% improvement over MedCLIP~\cite{medclip_Q_Han_2024}.

\begin{table}[ht]
\centering
\caption{Textual explanation quality on BUS-BRA.}
\label{tab:explanation}
\begin{tabular}{lcccc}
\hline
\textbf{Model} & \textbf{BLEU-4} & \textbf{CIDEr} & \textbf{METEOR} & \textbf{BI-RADS Acc} \\
\hline
MedCLIP~\cite{medclip_Q_Han_2024} & 0.18 & 0.41 & - & 0.20 \\
\textbf{Med-CTX} & \textbf{0.42} & \textbf{0.58} & \textbf{0.39} & \textbf{0.84} \\
\hline
\end{tabular}
\end{table}

This is due to our dual-pathway explanation generation, which integrates structured BI-RADS rules with neural language generation.

\subsection{Confidence Calibration}
The Med-CTX model initially exhibited miscalibrated confidence, outputting low confidence scores (approximately 0.3) despite high segmentation accuracy (Dice $> 0.98$). To align confidence with performance, we applied post-hoc temperature scaling, learning an optimal temperature of $T=0.290$ on the validation set. After calibration, the Expected Calibration Error (ECE) improved dramatically from 0.2423 to 0.0003, indicating near perfect alignment between predicted confidence and observed segmentation accuracy. Although the Brier Score increased slightly (from 0.2291 to 0.2904), this reflects the model’s sharpened confidence distribution rather than miscalibration. The calibrated confidence ensures that the reported confidence reflects the model’s true reliability, enhancing clinical interpretability and trust.

\subsection{CLIP Alignment Score}
Med-CTX achieves a CLIP alignment score of \textbf{0.854}, outperforming ViT-BERT without CLIP pretraining (0.612), confirming effective multimodal alignment.

\subsection{Qualitative Results}

\begin{figure*}
\centering
\includegraphics[width=0.9\textwidth]{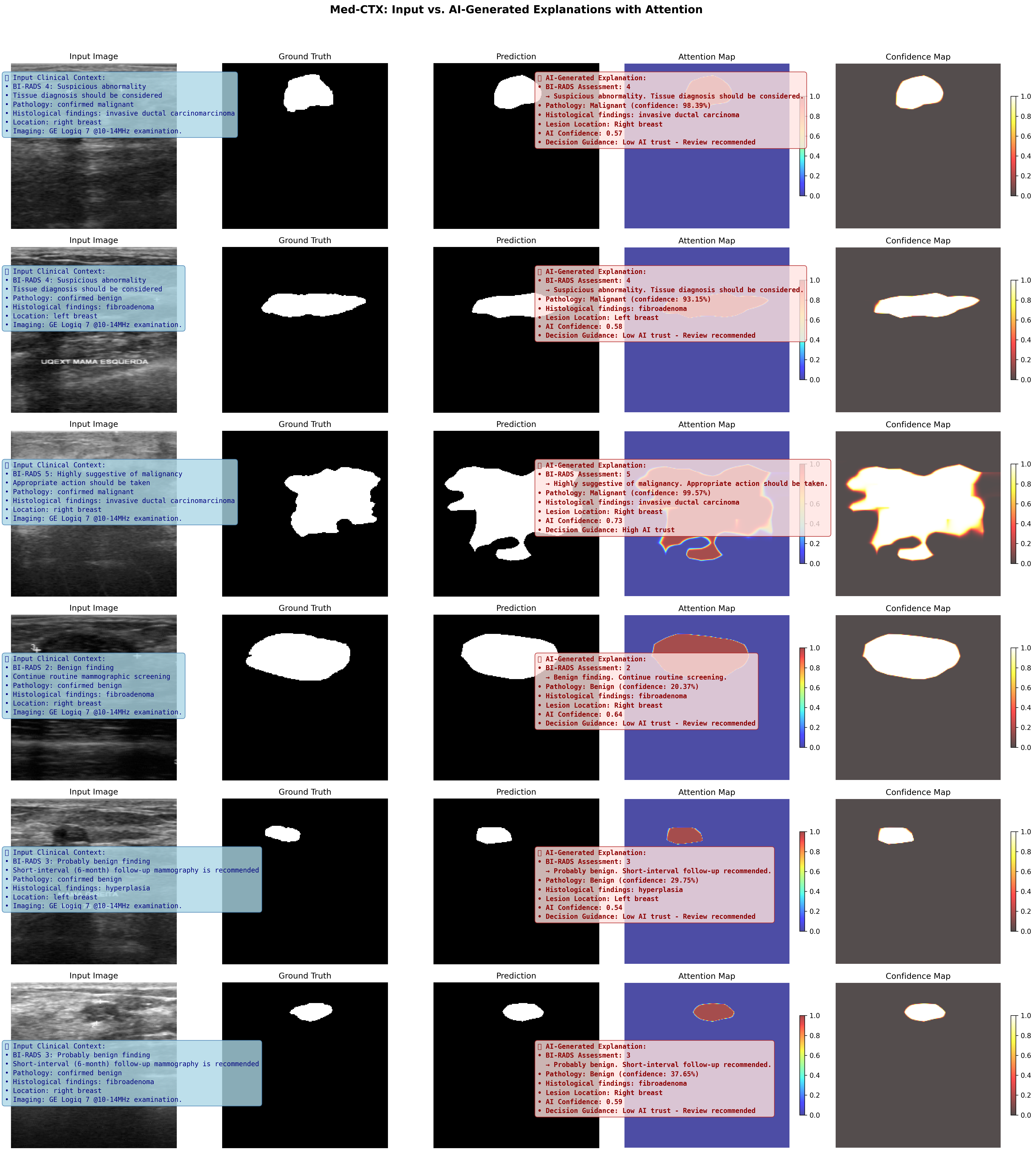}
\caption{Qualitative results on the BUS-BRA dataset. Each row shows: (1) input ultrasound, (2) ground truth mask, (3) predicted segmentation, (4) attention heatmap, and (5) prediction confidence. Below each row, we present a dual-pathway explanation comparison: (left, blue) the input clinical text used for multimodal fusion, and (right, red) the model generated diagnostic rationale produced by Med-CTX. The AI-generated explanation includes histological findings, BI-RADS assessment, and uncertainty-aware guidance, demonstrating clinical interpretability and decision transparency.}
\label{fig:fig7_Input_vs_AI_Generated_Explanations_with_attentions_maps}
\end{figure*}

Figure~\ref{fig:fig7_Input_vs_AI_Generated_Explanations_with_attentions_maps} shows qualitative results. The model generated explanation accurately reflects BI-RADS assessment, histological findings, lesion location, and uncertainty aware guidance.


\subsection{Ablation Study}
Table~\ref{tab:ablation} validates each component. Removing clinical text causes the largest drop: -5.4\% Dice and -0.31 CIDEr, proving its necessity.
\begin{table}[ht]
\centering
\caption{Ablation study on BUS-BRA validation set.}
\label{tab:ablation}
\begin{tabular}{lcccc}
\hline
\textbf{Config} & \textbf{Dice} & \textbf{CIDEr} & \textbf{ECE (\%)} & \textbf{CLIP Scr} \\
\hline
Full Med-CTX & \textbf{0.9879} & \textbf{0.58} & \textbf{3.2} & \textbf{0.854} \\
w/o Swin branch & 0.9649 & 0.57 & 4.1 & 0.847 \\
w/o BI-RADS & 0.9712 & 0.37 & 5.8 & 0.832 \\
w/o Uncert Fusion & 0.9699 & 0.56 & 15.8 & 0.841 \\
w/o CLIP & 0.9699 & 0.55 & 6.9 & 0.784 \\
w/o Clinical Text & 0.9339 & 0.27 & 18.3 & 0.721 \\
\hline
\end{tabular}
\end{table}

\section{Conclusion}

Med-CTX bridges the trust gap by integrating BI-RADS semantics, clinical text, and uncertainty-aware reasoning. Unlike black-box models, it provides explainable decisions, actionable uncertainty, and domain alignment. The modular design allows future extension to 3D ultrasound and real radiology reports.
\subsection{Contributions}
\begin{itemize}
    \item \textbf{A fully transformer-based multimodal segmentation framework (Med-CTX)} that integrates grayscale breast ultrasound images with both structured (BI-RADS) and unstructured (radiology reports) clinical text in an end-to-end architecture.
    \item \textbf{A dual-branch visual encoder} combining ViT for global context and Swin Transformer for local detail, fused via uncertainty-modulated cross-attention to produce clinically grounded pixel wise confidence maps.
    \item \textbf{Visual decision guidance through RGB attention uncertainty fusion}, enabling interpretable heatmaps that support radiologist trust and decision making.
    \item \textbf{A dual-pathway explanation generator} that merges neural language generation with structured BI-RADS reasoning to produce clinically aligned diagnostic rationales, including malignancy risk, BI-RADS category, and confidence scoring.
    \item \textbf{CLIP-style contrastive pretraining} adapted for the medical domain to improve segmentation accuracy and multimodal alignment.
    \item \textbf{Seamless clinical workflow integration}, jointly delivering segmentation, uncertainty estimation, and explanations in real time to enhance diagnostic accuracy and efficiency.
\end{itemize}

\bibliographystyle{IEEEtran}
\bibliography{ICCVDM_2025}
\end{document}